\documentclass{article}
\usepackage{PRIMEarxiv}

\usepackage[utf8]{inputenc}
\usepackage[T1]{fontenc}
\usepackage[hyphens]{url}
\usepackage{graphicx}
\usepackage{xcolor}
\usepackage{amsmath,amssymb}
\usepackage{array,booktabs}
\usepackage{algorithm,algorithmic}
\usepackage{float}
\usepackage[section]{placeins}
\usepackage{needspace}
\usepackage[numbers,sort&compress]{natbib}
\usepackage{microtype}
\usepackage[hidelinks]{hyperref}
\newcommand{\method}{CA$^3$C}
\newcommand{\methodfull}{Consequence-Aware Adaptive Action Chunking}
\newcommand{\policy}{\pi_\theta}
\newcommand{\wm}{f_\phi}
\newcommand{\R}{\mathbb{R}}
\DeclareUrlCommand\taskid{\urlstyle{rm}}

\definecolor{figblue}{RGB}{45,105,200}
\definecolor{figred}{RGB}{190,35,45}
\definecolor{figgreen}{RGB}{55,140,75}
\definecolor{figorange}{RGB}{220,115,55}
\definecolor{gainred}{RGB}{190,35,45}
\definecolor{lossgreen}{RGB}{55,140,75}
\newcommand{\tabgain}[1]{\textcolor{gainred}{(+#1)}}
\newcommand{\tabloss}[1]{\textcolor{lossgreen}{(-#1)}}
\newcommand{\tabblank}{\vphantom{(+0.00)}}

\hypersetup{pdfauthor={Yuyan Li, Yujia Wang, Yusong Huang, Junjie Yang, Yanggang Sheng, Ziyi Shi, Wenpeng Xu, Xiaoyang Zhou, Haoang Li, Hongliang Lu, Xinhu Zheng},pdftitle={Commit While Futures Agree: Consequence-Aware Adaptive Action Chunking for Robot Manipulation}}

\title{Commit While Futures Agree:\\Consequence-Aware Adaptive Action Chunking for Robot Manipulation}
\author{
  \textbf{Yuyan Li\textsuperscript{1,2} \quad Yujia Wang\textsuperscript{3} \quad Yusong Huang\textsuperscript{1,2} \quad Junjie Yang\textsuperscript{1,4} \quad Yanggang Sheng\textsuperscript{1}} \\
  \textbf{Ziyi Shi\textsuperscript{4} \quad Wenpeng Xu\textsuperscript{1} \quad Xiaoyang Zhou\textsuperscript{1,2} \quad Haoang Li\textsuperscript{1} \quad Hongliang Lu\textsuperscript{5} \quad Xinhu Zheng\textsuperscript{1,*}} \\[0.5em]
  \normalfont\small\textsuperscript{1}The Hong Kong University of Science and Technology (Guangzhou) \\
  \normalfont\small\textsuperscript{2}ANU Intelligent \qquad \textsuperscript{3}Victoria University of Wellington \\
  \normalfont\small\textsuperscript{4}The Hong Kong University of Science and Technology \\
  \normalfont\small\textsuperscript{5}Southern University of Science and Technology \\[0.4em]
  \normalfont\small\texttt{yli382@connect.hkust-gz.edu.cn} \\
  \normalfont\small\textsuperscript{*}Corresponding author: \texttt{xinhuzheng@hkust-gz.edu.cn}
}
\date{}

\begin{document}
\maketitle
\raggedbottom

\begin{abstract}Action-chunking policies predict multi-step control sequences, but a fundamental question remains: how much of a predicted action chunk should be committed before replanning? Existing systems typically execute a fixed-length prefix, implicitly assuming that the same execution horizon remains trustworthy across states. Some adaptive methods estimate this horizon from the similarity or stability of predicted actions. However, different actions may lead to the same successful outcome, whereas similar actions can produce different futures, suggesting that commitment should be determined by agreement among imagined futures rather than by similarity in action space. To this end, we propose \methodfull~(\method), an inference-time framework built on a simple principle: \textit{commit while imagined futures agree, and replan when they diverge}. Without modifying or retraining the base policy, \method{} uses an action-conditioned world model to imagine the future consequences of multiple candidate action chunks under the same sampling noise. Using these imagined consequences, we formulate execution-horizon estimation as a Bayesian change-point inference problem and select the execution candidate through future consensus. Across multiple simulation benchmarks and real-world robot manipulation tasks, \method{} consistently improves diverse action-chunking policies, achieving up to a 71.8\% relative reduction in failure rate over the corresponding base policies.

\end{abstract}

\section{Introduction}
Modern robot manipulation policies increasingly predict actions in chunks rather than one step at a time~\cite{kim2025openvlaoft,black2024pi0,shi2025pi05}.
This way captures richer temporal structure, reduces myopic switching, and amortizes expensive inference~\cite{zhao2023act,chi2023diffusion}.
However, one fundamental question remains: after predicting an action chunk, how much of it should the robot execute?

Most existing systems answer this question with a fixed execution prefix, implicitly assuming that predictions remain reliable for as long as they can be generated.
Prediction capability, however, does not imply execution reliability.
In free space, a policy may safely support long open-loop execution, whereas near grasping, contact, or precise placement, only a short execution prefix may remain reliable.
This discrepancy becomes even more pronounced under distribution shift, where an unfamiliar object pose or scene layout may invalidate an otherwise reliable open-loop plan much earlier than expected.
Conversely, frequent replanning can interrupt reliable execution, increase inference cost, and reduce motion consistency across independently generated action chunks.
Taken together, these limitations suggest that execution should be adaptive rather than governed by a fixed horizon.

\begin{figure}[!t]
    \centering
    \includegraphics[width=0.8\linewidth]{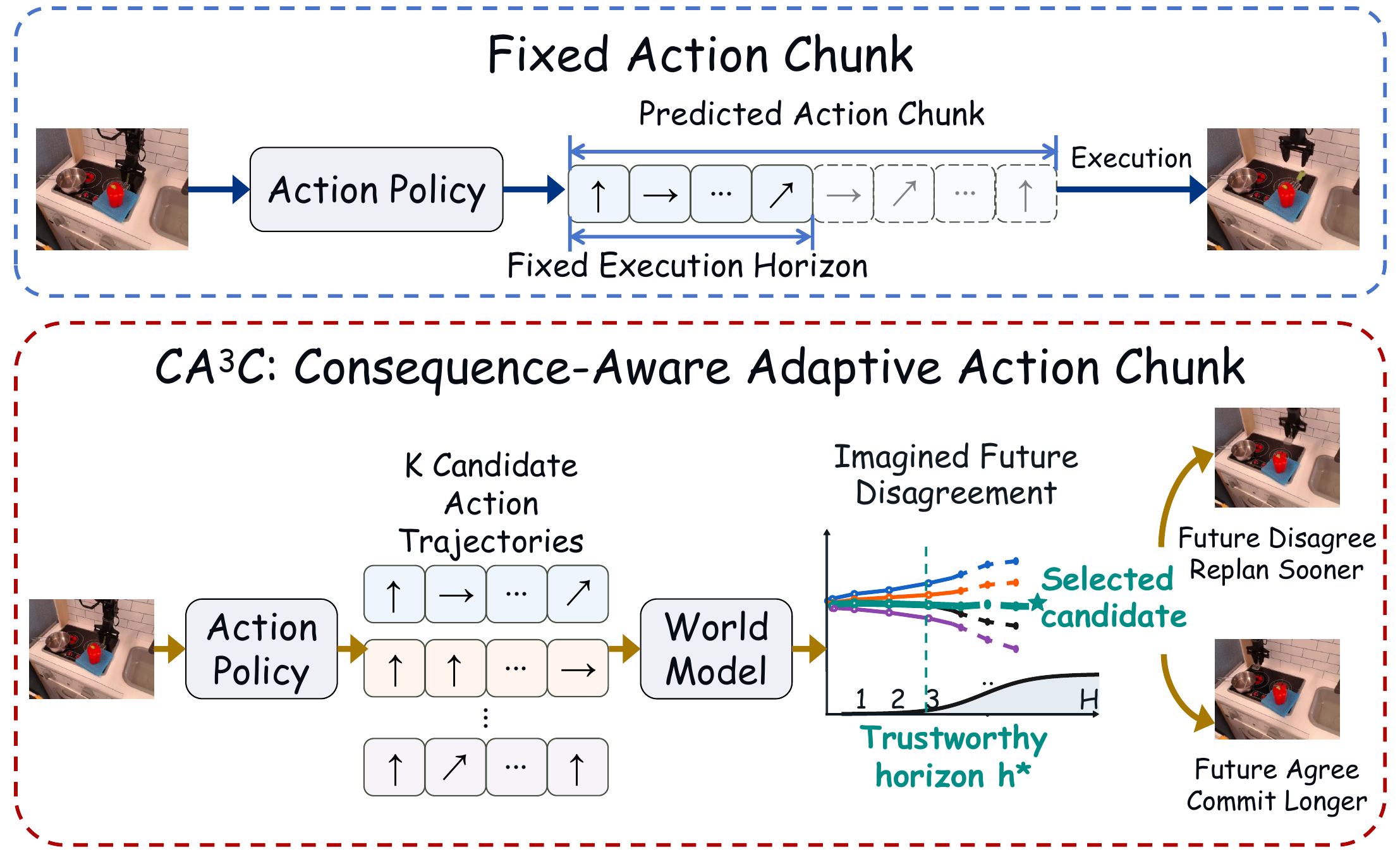}
    \caption{Fixed action chunking executes a preset prefix at every decision step. In contrast, \method{} compares the imagined consequences of multiple candidates, commits longer while their futures agree, replans sooner when they diverge, and executes the candidate closest to future consensus.}
    \label{fig:ca2c-teaser}
\end{figure}

Recent work makes execution adaptive using action entropy~\cite{liang2026aac}, agreement across prediction horizons~\cite{jing2026moh}, or stability along a flow policy's denoising trajectory~\cite{feng2026dvac}.
These methods recognize the importance of state-dependent commitment, but their confidence estimates are derived solely from the policy's action generation process.
Action ambiguity, however, does not necessarily imply execution risk.
Different action trajectories may produce the same successful outcome, while nearly identical actions near contact can instead lead to entirely different task outcomes.
Consequently, confidence measured in action space cannot directly determine whether the resulting future remains consistent and task-safe.
Taken together, the key question is not whether actions appear reliable, but whether the futures they induce remain consistent.

In this study, we propose \methodfull{} (\method), an inference-time framework for consequence-aware commitment.
Rather than asking whether sampled actions look alike, we ask whether their predicted effects remain mutually consistent.
Specifically, we introduce world models as an imagined-future consequence comparator.
At each replanning point, \method{} samples candidate action chunks and predicts their future latent trajectories while holding world-model sampling noise fixed.
It then measures candidate-induced disagreement relative to the total predicted scene change.
A Bayesian change-point test determines whether, and where, this disagreement begins to rise faster; no task-specific uncertainty threshold is introduced.
Finally, \method{} executes the candidate closest to the future consensus up to the selected boundary.
The resulting principle is concise: commit while imagined futures agree, and replan when they diverge.

Our contributions are summarized as follows:
\begin{itemize}
    \item We use an action-conditioned world model to select the execution length of each action chunk by locating where candidate-conditioned imagined futures begin to diverge.
    \item We use the same imagined futures to select the candidate closest to cross-candidate consensus, yielding a representative action sequence for execution.
    \item Across LIBERO, LIBERO-Plus, RoboCasa, and real-world AgiBot G2 tasks, \method{} consistently outperforms the corresponding base policies and strong baselines.
\end{itemize}

\section{Related Work}

\subsection{Adaptive Action Chunking}

Most action-chunking policies use a fixed execution horizon: they predict a multi-step sequence and execute the same preset number of actions before each replanning step~\cite{zhao2023act,chi2023diffusion,black2024pi0,shi2025pi05}.
Fixed-horizon execution is widely combined with temporal ensembling, guided candidate decoding, and asynchronous chunk stitching to improve temporal consistency or tolerate inference latency~\cite{zhao2023act,liu2025bid,black2025rtc}.
Recent research instead makes commitment state-dependent, using signals derived from policy outputs or the generation process~\cite{so2025sgac,liang2026aac,jing2026moh,feng2026dvac}, learned horizon selectors optimized through task feedback~\cite{zhao2026dehp}, or verification between predicted and subsequently observed futures~\cite{wang2026ffdc}.
\method{} likewise adapts commitment online, but determines the execution boundary before acting from agreement among the action-conditioned consequences of multiple candidate chunks, rather than from action-space confidence or post-execution prediction--observation discrepancies.

\subsection{World Models for Robot Manipulation}

World models predict how observations or latent states evolve under actions, enabling imagined-future reasoning in robot manipulation~\cite{ha2018worldmodels,hafner2023dreamerv3,zhu2024irasim,zhu2025uwm}.
Existing work uses such predictions primarily for visual planning and goal-conditioned control~\cite{finn2017deep,ebert2018visual,du2023unipi,zhou2024robodreamer,zhen2024threedvla}, policy or representation learning in imagination~\cite{hafner2023dreamerv3,wu2022daydreamer,zhu2025uwm,cen2025worldvla,zheng2025flare}, and action-conditioned simulation for policy evaluation, improvement, or data generation~\cite{zhu2024irasim,guo2025ctrlworld}.
Thus, imagined futures generally serve as plans, learning signals, or surrogate environments.
\method{} instead uses the world model as an execution-time consequence comparator for policy-proposed chunks: disagreement among candidate-conditioned futures determines how long to commit, while consensus identifies which candidate to execute.
To our knowledge, this use of a world model to jointly guide action-chunk truncation and candidate selection has not been studied previously.

\setcounter{secnumdepth}{2}
\section{Method}

\subsection{Preliminaries}

Our framework builds on two established model classes: an action-chunking policy that models a distribution over future controls, and an action-conditioned world model that models the future consequences of those controls.

\paragraph{Action-Chunking Policies.}
At decision time $t$, let $c_t$ denote the information available to a policy, such as observations, robot states, or task context.
An action-chunking policy models a conditional distribution over a length-$H$ action sequence~\cite{zhao2023act,chi2023diffusion},
\begin{equation}
A_t
=
(a_t,a_{t+1},\ldots,a_{t+H-1})
\sim
\policy(\cdot\mid c_t),
\end{equation}
where $a_{t+k}\in\R^{d_a}$ is a low-level control and $H$ is the prediction horizon.
We do not assume a particular policy architecture or input modality; the only requirement is that the policy can produce multiple plausible action chunks under the same condition.
In the standard action-chunk execution mechanism, the policy predicts $H$ actions but the controller executes only the first $h\leq H$ actions before observing the environment and replanning.
Thus, $H$ is the prediction horizon, whereas $h$ is the execution horizon; existing controllers commonly keep $h$ fixed throughout deployment~\cite{black2025rtc,zhao2026dehp}.

\begin{figure}[t]
    \centering
    \includegraphics[width=\textwidth]{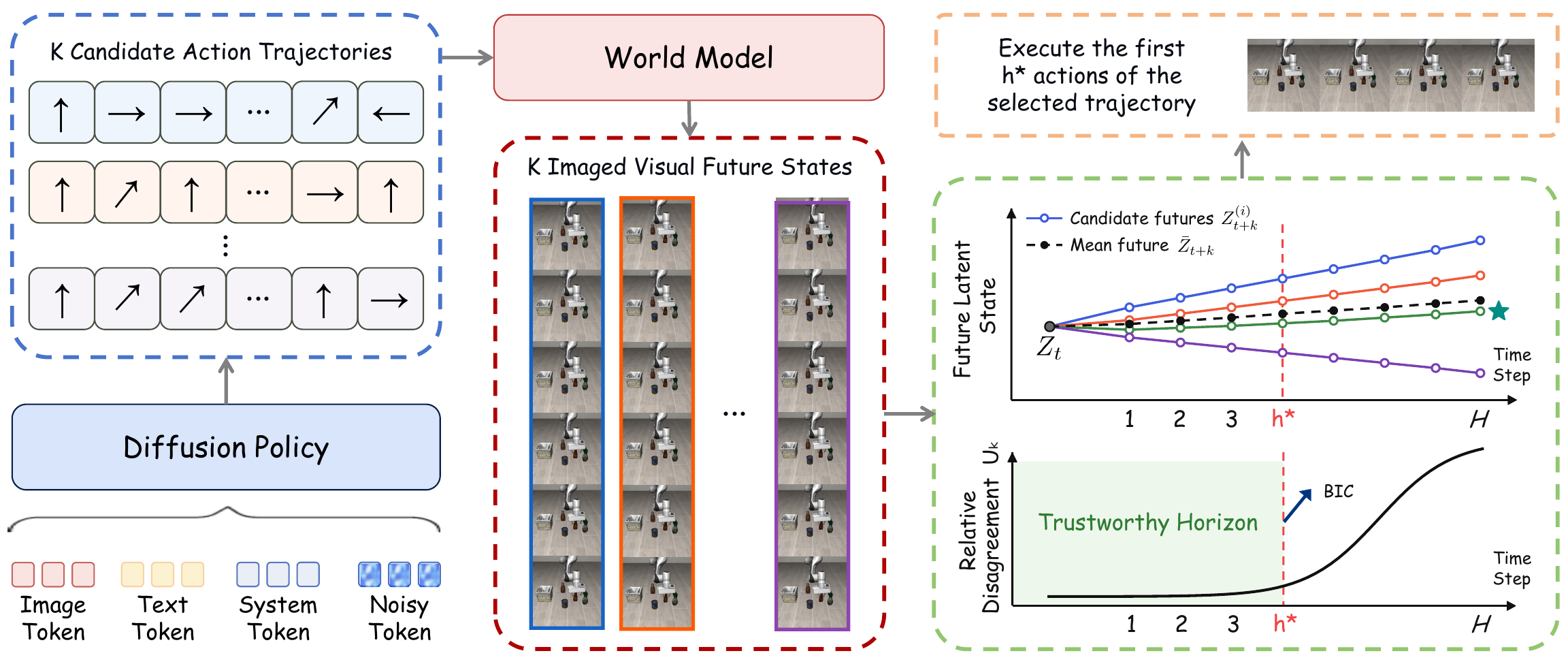}
    \caption{Overview of \methodfull{}.
    \textcolor{figblue}{\textbf{(a)}} The diffusion policy samples $K$ candidate action chunks.
    \textcolor{figred}{\textbf{(b)}} The action-conditioned world model imagines their visual future states under shared sampling noise.
    \textcolor{figgreen}{\textbf{(c)}} \method{} detects where their future latent features begin to disagree, determines the trustworthy horizon $h^\star$, and selects the candidate closest to future consensus.
    \textcolor{figorange}{\textbf{(d)}} The robot executes the first $h^\star$ actions of the selected chunk before replanning.}
    \label{fig:ca2c-framework}
\end{figure}

\paragraph{Action-Conditioned World Models.}
An action-conditioned world model predicts how the environment evolves under a proposed action chunk.
Given the current or historical context $x_{\leq t}$ and $A_t$, it models a distribution over a future trajectory~\cite{ha2018worldmodels,zhu2024irasim},
\begin{equation}
Y_{t+1:t+T}
\sim
p_\phi(\cdot\mid x_{\leq t},A_t).
\end{equation}
The future variable $Y$ may represent observations, latent states, or another task-relevant state representation.
The defining capability is to associate an action sequence with its time-aligned future consequences; no particular world-model architecture or generation procedure is assumed.

\subsection{Trustworthy Horizon Estimation}

Figure~\ref{fig:ca2c-framework} summarizes \method{}.
Its purpose is to estimate a trustworthy execution prefix at every replanning point; neither the policy nor the world model is updated.

At replanning time $t$, the policy samples $K$ candidate action chunks $\{A_t^{(i)}\}_{i=1}^K$ under the shared context $c_t=(o_t,q_t,\ell)$.
\method{} jointly selects a state-dependent horizon $h_t^\star\in\{h_{\min},\ldots,h_{\max}\}$ and a candidate $i_t^\star$, estimating how long their consequences remain mutually consistent rather than how many actions the policy can predict.

To estimate this horizon, an encoder maps the current observation to an anchor latent $z_t=E(o_t)$.
For each action candidate, the world model predicts
\begin{equation}
Z_t^{(i)}=\wm(z_t,A_t^{(i)},A_{<t})
=\{z_{t+1}^{(i)},\ldots,z_{t+T}^{(i)}\},
\end{equation}
where $A_{<t}$ is the action history required by the model.
We compare the first $T'=\min(T,H,h_{\max})$ future steps.
When $\wm$ is stochastic, independently sampled futures confound two effects: variation caused by candidate actions and variation caused by the predictor itself.
We therefore reuse the same initial world-model noise for all candidates within a replan, i.e., $\epsilon_t^{(1)}=\cdots=\epsilon_t^{(K)}$.
The predicted trajectories are consequently paired imagined futures: they share the current state and sampling realization, and differ in the proposed action chunk.

Given these controlled futures, at future step $k$, let the candidate mean be
\begin{equation}
\bar z_{t+k}=\frac{1}{K}\sum_{i=1}^{K}z_{t+k}^{(i)}.
\end{equation}
We define absolute candidate disagreement over latent dimension $d$ as
\begin{equation}
D_k=\frac{1}{Kd}\sum_{i=1}^{K}
\left\|z_{t+k}^{(i)}-\bar z_{t+k}\right\|_2^2.
\end{equation}
Raw disagreement varies with latent scale and with the amount of scene motion.
A large value can therefore reflect shared task progress rather than uncertainty about its outcome.
We normalize it by total predicted motion from the anchor,
\begin{equation}
M_k=\frac{1}{Kd}\sum_{i=1}^{K}
\left\|z_{t+k}^{(i)}-z_t\right\|_2^2,
\end{equation}
and compute
\begin{equation}
U_k=\frac{D_k}{M_k+\varepsilon}.
\label{eq:relative-disagreement}
\end{equation}
The variance decomposition
\begin{equation}
M_k=\frac{1}{d}\left\|\bar z_{t+k}-z_t\right\|_2^2+D_k
\end{equation}
shows that $U_k$ measures the fraction of predicted future change attributable to candidate-dependent divergence rather than shared progress.
We reduce isolated temporal noise with a width-three symmetric mean,
\begin{equation}
\widetilde U_k=\frac{1}{|\mathcal N_k|}
\sum_{j\in\mathcal N_k}U_j,
\quad
\mathcal N_k=\{k-1,k,k+1\}\cap\{1,\ldots,T'\}.
\end{equation}

We then convert the resulting disagreement curve into a commitment boundary.
The trustworthy horizon ends when consequence disagreement changes from a stable trend to a faster increase.
An absolute threshold is ill-suited to this decision because latent scales and scene dynamics vary across tasks.
Instead, we compare two explanations of the smoothed curve $y_k=\widetilde U_k$.
The null model is a single line,
\begin{equation}
f_0(k)=a+bk,
\quad
\mathrm{RSS}_0=\min_{a,b}\sum_{k=1}^{T'}(y_k-f_0(k))^2.
\end{equation}
For every $h\in\{h_{\min},\ldots,T'-1\}$, the alternative is a continuous hinge model,
\begin{equation}
f_h(k)=a+bk+c(k-h)_+,\qquad c\geq0,
\end{equation}
with residual $\mathrm{RSS}_1(h)$.
The nonnegative slope increment prevents an ordinary downward bend from being interpreted as increasing risk.
We score the models with the Bayesian information criterion~\cite{schwarz1978bic},
\begin{align}
\mathrm{BIC}_0 &=T'\log\frac{\mathrm{RSS}_0+\varepsilon}{T'}+2\log T',\\
\mathrm{BIC}_1(h)&=T'\log\frac{\mathrm{RSS}_1(h)+\varepsilon}{T'}+4\log T'.
\end{align}
Let $\widehat h=\arg\min_h\mathrm{BIC}_1(h)$.
The commitment horizon is
\begin{equation}
h_t^\star=
\begin{cases}
\widehat h,&\mathrm{BIC}_1(\widehat h)<\mathrm{BIC}_0,\\
T',&\text{otherwise.}
\end{cases}
\label{eq:horizon}
\end{equation}
Thus, a shorter prefix is selected only when the evidence for a change point justifies the additional model complexity.
If a single trend explains the curve, the imagined futures provide no supported boundary and \method{} commits to the longest valid prefix.

\subsection{Future-Consensus Candidate Selection}

Horizon selection determines \emph{when} to replan; future consensus determines \emph{which} candidate to execute.
For candidate $i$, define its consensus cost over the selected prefix as
\begin{equation}
C_i(h_t^\star)=\frac{1}{(K-1)h_t^\star d}
\sum_{j\neq i}\sum_{k=1}^{h_t^\star}
\left\|z_{t+k}^{(i)}-z_{t+k}^{(j)}\right\|_2^2.
\end{equation}
We select $i_t^\star=\arg\min_i C_i(h_t^\star)$ and execute the first $h_t^\star$ actions of $A_t^{(i_t^\star)}$.
For squared Euclidean distance, this is equivalent to choosing the candidate closest to the mean imagined trajectory.
The criterion favors a representative consequence rather than an extreme future and introduces no additional threshold.

\begin{algorithm}[H]
\caption{Consequence-aware adaptive commitment}
\label{alg:ca2c}
\begin{algorithmic}[1]
\REQUIRE Frozen policy $\policy$, world model $\wm$, bounds $h_{\min},h_{\max}$
\WHILE{the episode is not terminated}
    \STATE Sample $K$ candidate action chunks from $\policy$
    \STATE Imagine their future latents with shared world-model noise
    \STATE Compute $D_k$, $M_k$, and $U_k$ using Eq.~\ref{eq:relative-disagreement}
    \STATE Smooth $U_{1:T'}$ with a width-three temporal mean
    \STATE Select $h_t^\star$ by BIC model selection in Eq.~\ref{eq:horizon}
    \STATE Select the minimum-cost future-consensus candidate $i_t^\star$
    \STATE Execute the first $h_t^\star$ actions of $A_t^{(i_t^\star)}$
\ENDWHILE
\end{algorithmic}
\end{algorithm}

\section{Experiments}

We evaluate \methodfull{} (\method) across simulation and real-world tasks, including distribution shifts, component ablations, qualitative diagnostics, and cross-policy scalability.

\subsection{Experimental Setup}

We evaluate on LIBERO~\cite{liu2023libero} under its standard in-distribution (ID) setting and an out-of-distribution (OOD) setting with object initialization ranges enlarged by 3--9\,cm, 24 LIBERO-Plus~\cite{fei2025liberoplus} tasks, and 24 RoboCasa Kitchen tasks~\cite{nasiriany2024robocasa}.
Detailed task lists and perturbation settings are provided in the supplementary material.

We evaluate released benchmark-fine-tuned checkpoints of three representative generative action-chunking policies: GR00T N1.7~\cite{bjorck2025gr00t}, $\pi_0$~\cite{black2024pi0}, and $\pi_{0.5}$~\cite{shi2025pi05}.
Our action-conditioned world model is adapted from IRASim~\cite{zhu2024irasim} and fine-tuned for 60k steps on each of LIBERO and RoboCasa; further implementation details are provided in the supplementary material.

\subsection{Quantitative Results}

\paragraph{Results on LIBERO ID and OOD.}
Table~\ref{tab:libero-id-ood} reports results on standard LIBERO and its position-shifted OOD setting.
Across all three policies, \method{} yields 3.5--17.1\% relative reductions in average OOD failure rate.
In the ID setting, where baseline performance is already high, it preserves GR00T's average performance and improves both $\pi$ policies.

\begin{table}[!htbp]
\centering
{\small
\setlength{\tabcolsep}{2pt}
\begin{tabular*}{\textwidth}{@{\extracolsep{\fill}}lcccc!{\hspace{5pt}\vrule width \arrayrulewidth\hspace{-5pt}}c!{\hspace{4.5pt}\vrule width \arrayrulewidth\hspace{-4.5pt}}cccc!{\hspace{5pt}\vrule width \arrayrulewidth\hspace{-5pt}}c}
\toprule
& \multicolumn{5}{c!{\hspace{4.5pt}\vrule width \arrayrulewidth\hspace{-4.5pt}}}{ID} & \multicolumn{5}{c}{OOD}\\
\cmidrule(lr){2-6}\cmidrule(lr){7-11}
Method & Spatial & Object & Goal & L-10 & Avg. & Spatial & Object & Goal & L-10 & Avg.\\
\midrule
GR00T & 99.00 & 98.00 & 98.00 & 95.00 & 97.50 & 47.00 & 50.00 & 42.00 & 63.00 & 50.50\\
\shortstack[l]{\textbf{GR00T + \method{}}\\\tabblank} &
\shortstack{\textbf{98.00}\\\tabloss{1.00}} &
\shortstack{\textbf{98.00}\\\tabblank} &
\shortstack{\textbf{98.00}\\\tabblank} &
\shortstack{\textbf{96.00}\\\tabgain{1.00}} &
\shortstack{\textbf{97.50}\\\tabblank} &
\shortstack{\textbf{50.00}\\\tabgain{3.00}} &
\shortstack{\textbf{52.00}\\\tabgain{2.00}} &
\shortstack{\textbf{42.00}\\\tabblank} &
\shortstack{\textbf{65.00}\\\tabgain{2.00}} &
\shortstack{\textbf{52.25}\\\tabgain{1.75}}\\[4pt]
$\pi_0$ & 95.00 & 99.00 & 97.00 & 77.00 & 92.00 & 48.00 & 67.00 & 48.00 & 61.00 & 56.00\\
\shortstack[l]{\textbf{$\pi_{0}$ + \method{}}\\\tabblank} &
\shortstack{\textbf{94.00}\\\tabloss{1.00}} &
\shortstack{\textbf{99.00}\\\tabblank} &
\shortstack{\textbf{95.50}\\\tabloss{1.50}} &
\shortstack{\textbf{87.50}\\\tabgain{10.50}} &
\shortstack{\textbf{94.00}\\\tabgain{2.00}} &
\shortstack{\textbf{48.00}\\\tabblank} &
\shortstack{\textbf{77.00}\\\tabgain{10.00}} &
\shortstack{\textbf{51.00}\\\tabgain{3.00}} &
\shortstack{\textbf{65.00}\\\tabgain{4.00}} &
\shortstack{\textbf{60.25}\\\tabgain{4.25}}\\[4pt]
$\pi_{0.5}$ & 97.00 & 99.00 & 96.00 & 95.00 & 96.75 & 74.00 & 87.00 & 76.00 & 78.50 & 78.88\\
\shortstack[l]{\textbf{$\pi_{0.5}$ + \method{}}\\\tabblank} &
\shortstack{\textbf{99.00}\\\tabgain{2.00}} &
\shortstack{\textbf{99.00}\\\tabblank} &
\shortstack{\textbf{98.00}\\\tabgain{2.00}} &
\shortstack{\textbf{95.00}\\\tabblank} &
\shortstack{\textbf{97.75}\\\tabgain{1.00}} &
\shortstack{\textbf{76.00}\\\tabgain{2.00}} &
\shortstack{\textbf{90.00}\\\tabgain{3.00}} &
\shortstack{\textbf{80.00}\\\tabgain{4.00}} &
\shortstack{\textbf{84.00}\\\tabgain{5.50}} &
\shortstack{\textbf{82.50}\\\tabgain{3.62}}\\
\bottomrule
\end{tabular*}
}
\caption{Success rates (\%) on LIBERO ID and OOD. Baselines use regular type and rows augmented with \method{} are \textbf{bold}. Parentheses show absolute percentage-point changes from each baseline (\textcolor{gainred}{red}: increase; \textcolor{lossgreen}{green}: decrease); unchanged values are omitted. The same convention applies to subsequent main-result tables.}
\label{tab:libero-id-ood}
\end{table}

\paragraph{Results on LIBERO-Plus.}
Table~\ref{tab:liberoplus} shows that \method{} improves all three policies across all four suites, yielding relative reductions in average failure rate of 61.3\%, 30.5\%, and 45.4\% for GR00T, $\pi_0$, and $\pi_{0.5}$, respectively.
These gains span substantially different baseline strengths, while the suite-level relative failure-rate reduction reaches 71.8\% on LIBERO-Object for $\pi_{0.5}$.

\begin{table}[H]
\centering
{\small
\setlength{\tabcolsep}{2pt}
\begin{tabular*}{\columnwidth}{@{\extracolsep{\fill}}lcccc|c}
\toprule
Method & Spatial & Object & Goal & L-10 & Avg.\\
\midrule
GR00T & 47.5 & 48.3 & 45.8 & 41.7 & 45.8\\
\shortstack[l]{\textbf{GR00T + \method{}}\\\tabblank} &
\shortstack{\textbf{86.7}\\\tabgain{39.2}} &
\shortstack{\textbf{80.8}\\\tabgain{32.5}} &
\shortstack{\textbf{68.3}\\\tabgain{22.5}} &
\shortstack{\textbf{80.0}\\\tabgain{38.3}} &
\shortstack{\textbf{79.0}\\\tabgain{33.2}}\\[4pt]
$\pi_0$ & 75.0 & 60.8 & 60.0 & 75.8 & 67.9\\
\shortstack[l]{\textbf{$\pi_{0}$ + \method{}}\\\tabblank} &
\shortstack{\textbf{75.8}\\\tabgain{0.8}} &
\shortstack{\textbf{80.8}\\\tabgain{20.0}} &
\shortstack{\textbf{60.8}\\\tabgain{0.8}} &
\shortstack{\textbf{93.3}\\\tabgain{17.5}} &
\shortstack{\textbf{77.7}\\\tabgain{9.8}}\\[4pt]
$\pi_{0.5}$ & 93.3 & 88.3 & 73.3 & 97.5 & 88.1\\
\shortstack[l]{\textbf{$\pi_{0.5}$ + \method{}}\\\tabblank} &
\shortstack{\textbf{96.7}\\\tabgain{3.4}} &
\shortstack{\textbf{96.7}\\\tabgain{8.4}} &
\shortstack{\textbf{80.8}\\\tabgain{7.5}} &
\shortstack{\textbf{100.0}\\\tabgain{2.5}} &
\shortstack{\textbf{93.5}\\\tabgain{5.4}}\\
\bottomrule
\end{tabular*}
}
\caption{Success rates (\%) on 24 LIBERO-Plus tasks. Each suite contains three Robot Initial States tasks and three Objects Layout tasks and is evaluated with 120 rollouts.}
\label{tab:liberoplus}
\end{table}

\paragraph{Results on RoboCasa Kitchen.}
Table~\ref{tab:robocasa} shows that \method{} improves all three policies, reducing the average failure rates of GR00T, $\pi_0$, and $\pi_{0.5}$ by 11.6\%, 4.6\%, and 7.0\% relative to their baselines.
Gains across all four task groups demonstrate transfer beyond LIBERO.

\begin{table}[H]
\centering
{\small
\setlength{\tabcolsep}{2pt}
\begin{tabular*}{\columnwidth}{@{\extracolsep{\fill}}lcccc|c}
\toprule
Method & Relocation & Container & Rotation & Button & Avg.\\
\midrule
GR00T & 54.0 & 80.8 & 77.0 & 96.7 & 70.8\\
\shortstack[l]{\textbf{GR00T + \method{}}\\\tabblank} &
\shortstack{\textbf{57.5}\\\tabgain{3.5}} &
\shortstack{\textbf{85.0}\\\tabgain{4.2}} &
\shortstack{\textbf{81.0}\\\tabgain{4.0}} &
\shortstack{\textbf{96.7}\\\tabblank} &
\shortstack{\textbf{74.2}\\\tabgain{3.4}}\\[4pt]
$\pi_0$ & 19.0 & 55.8 & 46.0 & 56.7 & 38.5\\
\shortstack[l]{\textbf{$\pi_{0}$ + \method{}}\\\tabblank} &
\shortstack{\textbf{21.0}\\\tabgain{2.0}} &
\shortstack{\textbf{58.3}\\\tabgain{2.5}} &
\shortstack{\textbf{49.0}\\\tabgain{3.0}} &
\shortstack{\textbf{61.7}\\\tabgain{5.0}} &
\shortstack{\textbf{41.3}\\\tabgain{2.8}}\\[4pt]
$\pi_{0.5}$ & 44.0 & 59.2 & 59.0 & 60.0 & 52.9\\
\shortstack[l]{\textbf{$\pi_{0.5}$ + \method{}}\\\tabblank} &
\shortstack{\textbf{46.0}\\\tabgain{2.0}} &
\shortstack{\textbf{63.3}\\\tabgain{4.1}} &
\shortstack{\textbf{62.0}\\\tabgain{3.0}} &
\shortstack{\textbf{66.7}\\\tabgain{6.7}} &
\shortstack{\textbf{56.2}\\\tabgain{3.3}}\\
\bottomrule
\end{tabular*}
}
\caption{Success rates (\%) on the 24 RoboCasa Kitchen tasks. Relocation, Container, Rotation, and Button contain 10, 6, 5, and 3 tasks, respectively; each task is evaluated with 20 rollouts.}
\label{tab:robocasa}
\end{table}

\FloatBarrier
\subsection{Ablation Study}

\paragraph{Contribution of Each Component.}
We ablate \method{} on 24 LIBERO-Plus tasks using $\pi_{0.5}$.
Table~\ref{tab:component-ablation} compares the full method with a selection-only variant using the fixed baseline horizon $h=5$ and a horizon-only variant that always executes the primary candidate.
All other components and evaluation settings follow the main experiment.

\begin{table}[H]
\centering
{\small
\setlength{\tabcolsep}{2pt}
\begin{tabular*}{\columnwidth}{@{\extracolsep{\fill}}ccccccc}
\toprule
\shortstack{Adaptive\\Horizon} & \shortstack{Consensus\\Selection} & Spatial & Object & Goal & L-10 & Avg.\\
\midrule
 &  & 93.3 & \underline{88.3} & 73.3 & 97.5 & 88.1\\
 & $\checkmark$ & \underline{95.0} & 81.7 & 70.0 & \underline{98.3} & 86.3\\
$\checkmark$ &  & \underline{95.0} & \textbf{96.7} & \underline{80.0} & 93.3 & \underline{91.3}\\
$\checkmark$ & $\checkmark$ & \textbf{96.7} & \textbf{96.7} & \textbf{80.8} & \textbf{100.0} & \textbf{93.5}\\
\bottomrule
\end{tabular*}
}
\caption{Component ablation on 24 LIBERO-Plus tasks. The best result in each column is shown in \textbf{bold}, and the second-best result is \underline{underlined}. Checkmarks indicate enabled components, and values are success rates (\%). All variants use the same rollout protocol as the main experiment.}
\label{tab:component-ablation}
\end{table}

Adaptive horizon estimation alone yields a 26.9\% relative reduction in failure rate, whereas consensus selection alone does not improve the fixed-horizon baseline.
The full method reaches a 45.4\% relative reduction, identifying adaptive horizon selection as the core mechanism and future-consensus selection as a complementary refinement.

\paragraph{Comparison with Fixed Commitment Horizons.}
Table~\ref{tab:fixed-horizon-sensitivity} compares $\pi_{0.5}$ with $h\in\{4,5,8,10\}$ while disabling adaptive horizon estimation and consensus selection, testing whether a better fixed horizon can reproduce the gain of \method{}.

\begin{table}[H]
\centering
{\small
\setlength{\tabcolsep}{2pt}
\begin{tabular*}{\columnwidth}{@{\extracolsep{\fill}}lccccc}
\toprule
Fixed Horizon & Spatial & Object & Goal & L-10 & Avg.\\
\midrule
$h=4$ & \underline{96.7} & 88.3 & \underline{78.3} & 93.3 & \underline{89.2}\\
$h=5$ & 90.0 & 90.0 & 73.3 & 95.0 & 87.1\\
$h=8$ & \textbf{98.3} & \underline{96.7} & \underline{78.3} & \textbf{98.3} & \textbf{92.9}\\
$h=10$ & 95.0 & \textbf{100.0} & \textbf{80.0} & \underline{96.7} & \textbf{92.9}\\
\bottomrule
\end{tabular*}
}
\caption{Fixed-horizon comparison for the $\pi_{0.5}$ baseline on LIBERO-Plus (60 rollouts per suite). Best and second-best results are \textbf{bold} and \underline{underlined}.}
\label{tab:fixed-horizon-sensitivity}
\end{table}

Although fixed $h=8$ and $h=10$ each reduce average failure by 45.0\% relative to $h=5$, neither is best across all suites.
This suite-dependent optimum shows that no single global horizon consistently recovers the benefit of state-dependent commitment.

\begin{figure}[!tb]
\centering
\includegraphics[width=\textwidth]{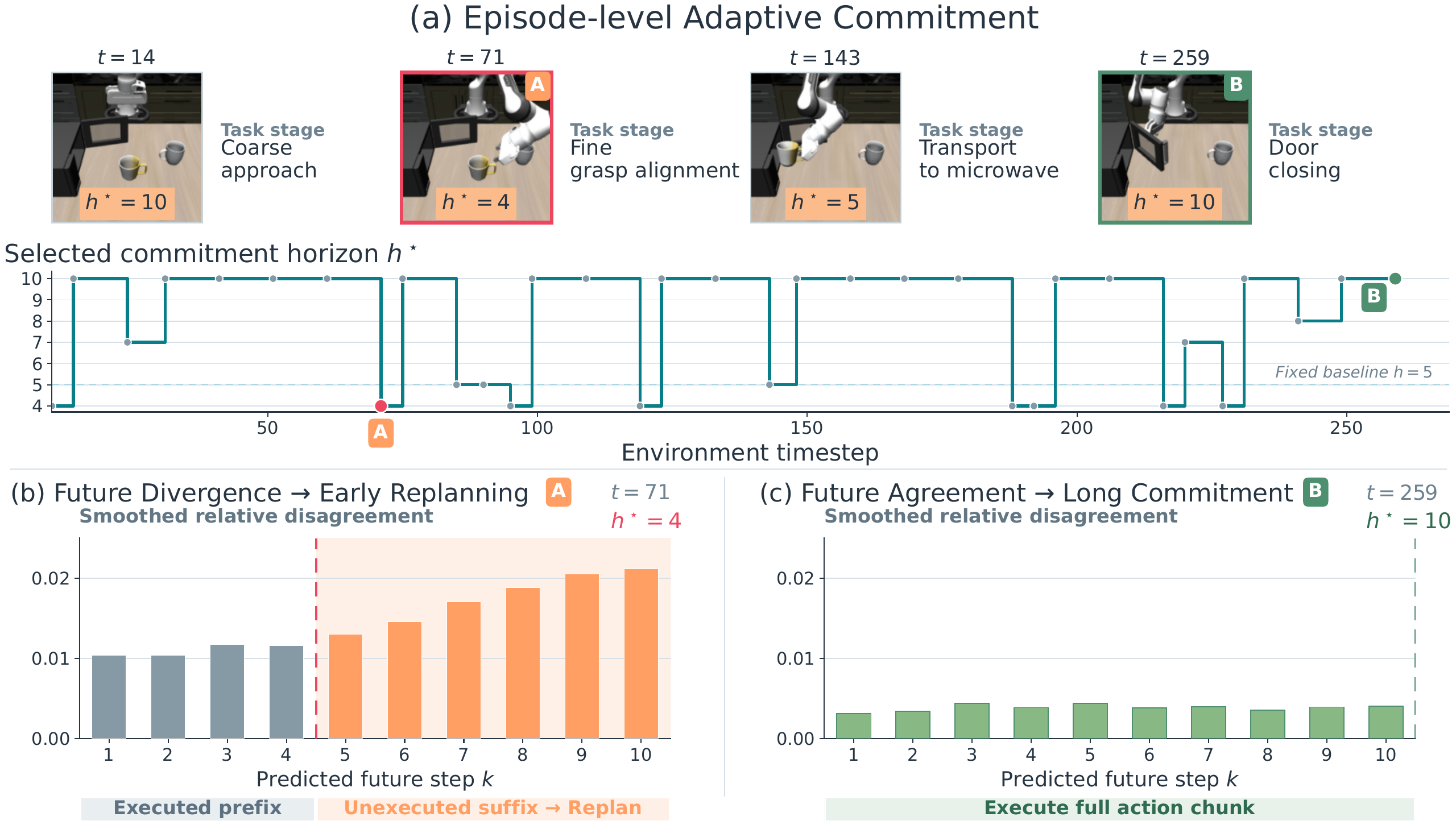}
\caption{Episode-level adaptive commitment for the instruction \emph{Put the yellow and white mug in the microwave and close it.} Panel~(a) traces the selected horizon $h^\star$ across the episode and shows representative task stages; the dashed line denotes the fixed $h=5$ baseline, and A/B mark the two decisions enlarged below. Panel~(b) shows that disagreement increases after future step 4 at A, so only the trustworthy prefix is executed before replanning. Panel~(c) shows consistently low disagreement at B, supporting execution of the full action chunk with $h^\star=10$.}
\label{fig:episode-adaptive-commitment}
\end{figure}

\subsection{Qualitative Results}

\paragraph{Episode-Level Adaptive Commitment.}
Figure~\ref{fig:episode-adaptive-commitment} visualizes the commitment decisions made by \method{} throughout a successful manipulation episode.
The selected horizon varies between 4 and 10 across coarse approach, fine grasp alignment, transport, and door closing, rather than collapsing to the fixed $h=5$ baseline.
At decision A ($t=71$), disagreement increases beyond future step 4, prompting early replanning after the trustworthy prefix; at decision B ($t=259$), consistently low disagreement supports executing the full chunk with $h^\star=10$.
Thus, the same policy alternates between reactive execution and long commitment according to the imagined consequences at each state.

\begin{figure}[!tb]
\centering
\includegraphics[width=\textwidth]{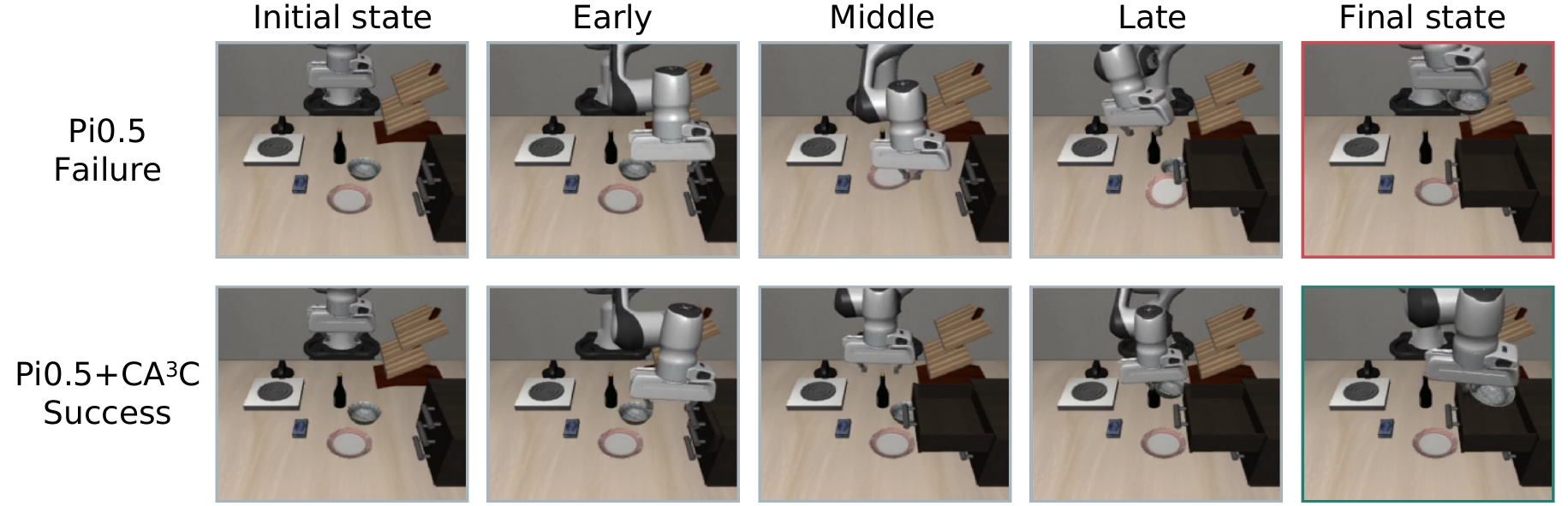}
\caption{Matched rollouts on LIBERO-Goal OOD for the instruction \emph{Open the top drawer and put the bowl inside.} The top and bottom rows show $\pi_{0.5}$ and $\pi_{0.5}$ + \method{}, respectively, from the same OOD initial state. Columns show representative execution stages, with red and green borders denoting the final failure and success outcomes.}
\label{fig:failure-to-success}
\end{figure}

\paragraph{Failure-to-Success Comparison.}
Figure~\ref{fig:failure-to-success} provides a matched rollout comparison that complements the aggregate quantitative results.
Although both policies make similar early progress in opening the drawer and manipulating the bowl, the base policy fails to complete the final placement, whereas the policy equipped with \method{} carries the execution through to task completion.
This contrast shows that the benefit of \method{} is reflected not only in higher success rates, but also in more reliable execution during the contact-sensitive final stage, where an inappropriate action commitment can otherwise turn substantial intermediate progress into task failure.

\begin{figure}[!htbp]
\centering
\includegraphics[width=0.65\linewidth]{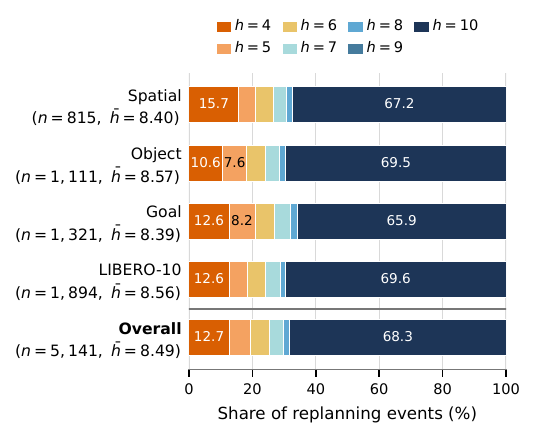}
\caption{Suite-level commitment-horizon distributions on LIBERO-Plus. Bars are normalized within each suite; labels show event counts, mean horizons, and frequent-horizon.}
\label{fig:suite-horizon-distribution}
\end{figure}

\paragraph{Suite-Level Distribution of Selected Horizons.}
Figure~\ref{fig:suite-horizon-distribution} aggregates 5,141 replanning events from the horizon-only variant.
Although $h^\star=10$ dominates across all suites, 31.7\% of decisions select a shorter horizon between 4 and 9, reflecting the alternation between stable transport phases and alignment- or contact-sensitive stages that require earlier feedback.
Spatial and Goal exhibit slightly lower mean horizons than Object and LIBERO-10, consistent with their greater sensitivity to spatial alignment and goal-conditioned interactions.
The estimator therefore neither collapses to the fixed $h=5$ baseline nor always executes the maximum horizon; it dynamically preserves long commitments in stable states and shortens them when the task demands tighter feedback.

\begin{table}[!htbp]
\centering
{\small
\setlength{\tabcolsep}{2pt}
\begin{tabular*}{\columnwidth}{@{\extracolsep{\fill}}lccc}
\toprule
Method & Block Separation & Parcel Sorting & Avg.\\
\midrule
ACOT & 78.0 & 56.0 & 67.0\\
\shortstack[l]{ACOT + \method{}\\\tabblank} &
\shortstack{82.0\\\tabgain{4.0}} &
\shortstack{62.0\\\tabgain{6.0}} &
\shortstack{72.0\\\tabgain{5.0}}\\
\bottomrule
\end{tabular*}
}
\caption{Success rates (\%) on real-world tasks using AgiBot G2. Each method is evaluated with 50 rollouts per task.}
\label{tab:real-world}
\end{table}

\subsection{Real-World Experiments}

\begin{figure}[!tb]
\centering
\includegraphics[width=\textwidth]{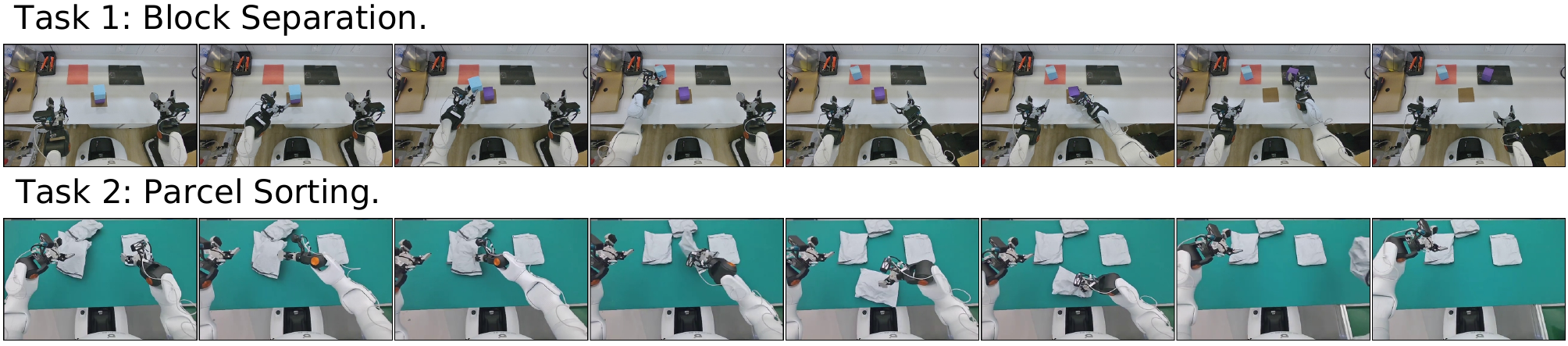}
\caption{Execution examples for real-world tasks using \method{}.}
\label{fig:real-world-execution}
\end{figure}

\paragraph{Setup and Tasks.}
We evaluate \method{} on an AgiBot G2 using ACOT~\cite{zhong2026acotvla} on two tasks: short-horizon \textbf{Block Separation}, which separates two stacked blocks into distinct target areas, and long-horizon \textbf{Parcel Sorting}, which transfers a parcel to a conveyor, flipping it first if its QR code faces downward.
Figure~\ref{fig:real-world-execution} shows representative execution sequences for both tasks.

\paragraph{Quantitative Results.}
Table~\ref{tab:real-world} shows that ACOT+\method{} raises average success from 67.0\% to 72.0\%, a 15.2\% relative reduction in failure rate.
The larger Parcel Sorting gain extends adaptive commitment to a fourth policy and a longer, condition-dependent real-world task.

\section{Discussion}

Consequence disagreement is neither a confidence score for a single prediction nor a calibrated estimate of model error; it measures how predicted outcomes vary across plausible actions, allowing the controller to commit while their consequences agree and replan when they diverge.
\method{} therefore requires a generative policy with meaningful candidate diversity and an action-sensitive world model: nearly deterministic policies offer little variation to compare, whereas action-insensitive or inaccurate world models can cause false agreement or divergence.
Imagined-future prediction also incurs inference cost, and structural choices such as candidate count, minimum horizon, and temporal smoothing remain despite BIC removing the need for an absolute uncertainty threshold.
Thus, \method{} is best suited to settings where diverse plausible chunks are available and the world model preserves their control-relevant consequences.

\clearpage
\begingroup
\setlength{\bibsep}{2pt}
\bibliographystyle{unsrtnat}
\bibliography{references}
\endgroup

\clearpage
\appendix
\section{Additional Method Details}\subsection{World-Model Architecture and Training}

\method{} uses an action-conditioned latent video diffusion model adapted from IRASim~\cite{zhu2024irasim}.
A frozen SDXL VAE encodes RGB observations into four-channel latents at one-eighth spatial resolution.
The denoiser is an IRASim-XL/2 diffusion transformer with 28 alternating spatial and temporal blocks, a hidden width of 1152, 16 attention heads, and a latent patch size of 2.
The current observation latent is kept clean as the conditioning anchor, while future latent positions are initialized with Gaussian noise.
Frame-aligned action embeddings and the diffusion timestep modulate the spatial blocks through adaptive layer normalization, while temporal attention propagates information across the prediction horizon.
For each candidate action chunk, the model predicts 15 future latents, which are directly used by \method{} without RGB decoding; decoded frames are produced only for visualization.

To improve sensitivity to action-induced state changes, training augments the standard diffusion objective with three auxiliary objectives.
Bidirectional visual prediction reconstructs four preceding frames while predicting the future around the clean anchor; action--state transition consistency uses a lightweight inverse-dynamics head to recover actions from adjacent predicted latents; and latent motion-aware prediction emphasizes changing regions while regularizing the static background.
These auxiliary targets, heads, and masks are used only during training and removed at inference.
We separately fine-tune benchmark-specific world models on LIBERO and RoboCasa action--observation sequences for 60k steps.
The resulting world models remain frozen throughout all \method{} experiments.

\subsection{Complete Inference Procedure}

At each replanning step, the frozen policy independently samples $K=3$ candidate action chunks under the same observation, robot state, task instruction, and action history.
The frozen world model predicts a future latent sequence for each candidate from the same current observation.
All candidates share the same initial diffusion noise within a replanning step, so their differences primarily reflect the consequences of the proposed actions.
We compare the first $T'=\min(T,H,h_{\max})$ valid future steps.

\method{} computes the relative disagreement $U_k$, applies a width-three temporal mean, and uses BIC to compare a single linear trend with a continuous hinge model whose slope can increase after a change point.
The detected boundary determines $h_t^\star$; if the hinge model is not preferred, the longest valid prefix is used.
Over this prefix, \method{} selects the candidate with the smallest mean squared latent distance to the others, executes its first $h_t^\star$ actions, and then replans from the new observation.
We use $h_{\min}=4$, $h_{\max}=15$, $\varepsilon=10^{-8}$, and ten world-model diffusion steps.
The available policy chunk limits the realized horizon to $h_t^\star\in\{4,\ldots,10\}$ in our experiments.

\subsection{Additional Derivations}

\paragraph{Interpretation of Relative Disagreement.}
Let $x_i=z_{t+k}^{(i)}-z_t$ denote the predicted latent change of candidate $i$, and let $\bar{x}=K^{-1}\sum_i x_i$.
Since $\sum_i(x_i-\bar{x})=0$, expanding $x_i=(x_i-\bar{x})+\bar{x}$ gives
\[
M_k
=
\frac{1}{d}\|\bar{x}\|_2^2
+
\frac{1}{Kd}\sum_{i=1}^{K}\|x_i-\bar{x}\|_2^2
=
\frac{1}{d}\|\bar{z}_{t+k}-z_t\|_2^2+D_k.
\]
Thus, $M_k$ decomposes into motion shared across candidates and candidate-dependent disagreement.
Because $D_k\leq M_k$, the normalized signal $U_k=D_k/(M_k+\varepsilon)$ satisfies $0\leq U_k<1$.
Shared task progress can therefore produce substantial motion without increasing $U_k$, whereas separation among candidate-conditioned futures raises the relative disagreement.

\paragraph{Future Consensus and the Mean Imagined Trajectory.}
Concatenate the predicted latents of candidate $i$ over the selected prefix as $x_i=[z_{t+1}^{(i)},\ldots,z_{t+h_t^\star}^{(i)}]$, and let $\bar{x}=K^{-1}\sum_i x_i$.
The pairwise-distance identity
\[
\sum_{j=1}^{K}\|x_i-x_j\|_2^2
=
K\|x_i-\bar{x}\|_2^2
+
\sum_{j=1}^{K}\|x_j-\bar{x}\|_2^2
\]
shows that the second term is independent of candidate $i$.
Minimizing the future-consensus cost is therefore equivalent to selecting the candidate closest to the mean imagined trajectory.
This favors a representative future without introducing an additional selection threshold.

\section{Detailed Experimental Protocols}

\subsection{Policy Checkpoints}

We keep all policy parameters frozen during evaluation and use the same checkpoint for each baseline and its \method{}-augmented variant.
For LIBERO and LIBERO-Plus, we use the GR00T N1.7 checkpoint \texttt{nvidia/GR00T-N1.7-LIBERO}, with suite-specific weights for LIBERO-Spatial, LIBERO-Object, LIBERO-Goal, and LIBERO-10.
For the $\pi$ policies, we use the official OpenPI checkpoints \texttt{pi0\_libero} and \texttt{pi05\_libero} for $\pi_0$ and $\pi_{0.5}$, respectively.
For RoboCasa, we use a RoboCasa-fine-tuned GR00T N1.7 checkpoint.
The $\pi_0$ and $\pi_{0.5}$ models are initialized from their official base checkpoints and fine-tuned for 30k steps, and the resulting checkpoints are used for evaluation.
The real-world experiments use task-specific ACOT checkpoints fine-tuned on AgiBot G2 demonstrations.

\subsection{LIBERO ID and OOD}

We evaluate the same 40 tasks from the four standard LIBERO suites~\cite{liu2023libero} under both the in-distribution (ID) and position-shifted out-of-distribution (OOD) settings.
The OOD setting enlarges the object-initialization ranges by 3--9 cm relative to the corresponding ID ranges, while retaining the original task instructions and success criteria.
Each suite is evaluated with 100 rollouts under each condition.

\paragraph{LIBERO-Spatial.}
\begin{enumerate}
\setlength{\itemsep}{0pt}
\item Pick up the black bowl between the plate and the ramekin and place it on the plate.
\item Pick up the black bowl next to the ramekin and place it on the plate.
\item Pick up the black bowl from table center and place it on the plate.
\item Pick up the black bowl on the cookie box and place it on the plate.
\item Pick up the black bowl in the top drawer of the wooden cabinet and place it on the plate.
\item Pick up the black bowl on the ramekin and place it on the plate.
\item Pick up the black bowl next to the cookie box and place it on the plate.
\item Pick up the black bowl on the stove and place it on the plate.
\item Pick up the black bowl next to the plate and place it on the plate.
\item Pick up the black bowl on the wooden cabinet and place it on the plate.
\end{enumerate}

\paragraph{LIBERO-Object.}
\begin{enumerate}
\setlength{\itemsep}{0pt}
\item Pick up the alphabet soup and place it in the basket.
\item Pick up the cream cheese and place it in the basket.
\item Pick up the salad dressing and place it in the basket.
\item Pick up the BBQ sauce and place it in the basket.
\item Pick up the ketchup and place it in the basket.
\item Pick up the tomato sauce and place it in the basket.
\item Pick up the butter and place it in the basket.
\item Pick up the milk and place it in the basket.
\item Pick up the chocolate pudding and place it in the basket.
\item Pick up the orange juice and place it in the basket.
\end{enumerate}

\paragraph{LIBERO-Goal.}
\begin{enumerate}
\setlength{\itemsep}{0pt}
\item Open the middle drawer of the cabinet.
\item Put the bowl on the stove.
\item Put the wine bottle on top of the cabinet.
\item Open the top drawer and put the bowl inside.
\item Put the bowl on top of the cabinet.
\item Push the plate to the front of the stove.
\item Put the cream cheese in the bowl.
\item Turn on the stove.
\item Put the bowl on the plate.
\item Put the wine bottle on the rack.
\end{enumerate}

\paragraph{LIBERO-10.}
\begin{enumerate}
\setlength{\itemsep}{0pt}
\item Put both the alphabet soup and the tomato sauce in the basket.
\item Put both the cream cheese box and the butter in the basket.
\item Turn on the stove and put the moka pot on it.
\item Put the black bowl in the bottom drawer of the cabinet and close it.
\item Put the white mug on the left plate and put the yellow and white mug on the right plate.
\item Pick up the book and place it in the back compartment of the caddy.
\item Put the white mug on the plate and put the chocolate pudding to the right of the plate.
\item Put both the alphabet soup and the cream cheese box in the basket.
\item Put both moka pots on the stove.
\item Put the yellow and white mug in the microwave and close it.
\end{enumerate}

\subsection{LIBERO-Plus}

We evaluate 24 LIBERO-Plus tasks~\cite{fei2025liberoplus} spanning the same four suites.
Each suite contains three Robot Initial States tasks and three Objects Layout tasks, listed below by their exact task identifiers.
Each task is evaluated with 20 rollouts, yielding 120 rollouts per suite.

\paragraph{LIBERO-Spatial.}
\textit{Robot Initial States}
\begin{enumerate}
\setlength{\itemsep}{0pt}
\item \taskid{pick_up_the_black_bowl_between_the_plate_and_the_ramekin_and_place_it_on_the_plate_view_0_0_100_0_0_initstate_1}
\item \taskid{pick_up_the_black_bowl_from_table_center_and_place_it_on_the_plate_view_0_0_100_0_0_initstate_103}
\item \taskid{pick_up_the_black_bowl_in_the_top_drawer_of_the_wooden_cabinet_and_place_it_on_the_plate_view_0_0_100_0_0_initstate_135}
\end{enumerate}
\Needspace{4\baselineskip}
\textit{Objects Layout}
\begin{enumerate}
\setlength{\itemsep}{0pt}
\item \taskid{pick_up_the_black_bowl_between_the_plate_and_the_ramekin_and_place_it_on_the_plate_level5_sample3}
\item \taskid{pick_up_the_black_bowl_from_table_center_and_place_it_on_the_plate_level1_sample1}
\item \taskid{pick_up_the_black_bowl_in_the_top_drawer_of_the_wooden_cabinet_and_place_it_on_the_plate_level1_sample1}
\end{enumerate}

\paragraph{LIBERO-Object.}
\textit{Robot Initial States}
\begin{enumerate}
\setlength{\itemsep}{0pt}
\item \taskid{pick_up_the_alphabet_soup_and_place_it_in_the_basket_view_0_0_100_0_0_initstate_101}
\item \taskid{pick_up_the_bbq_sauce_and_place_it_in_the_basket_view_0_0_100_0_0_initstate_224}
\item \taskid{pick_up_the_butter_and_place_it_in_the_basket_view_0_0_100_0_0_initstate_117}
\end{enumerate}
\Needspace{4\baselineskip}
\textit{Objects Layout}
\begin{enumerate}
\setlength{\itemsep}{0pt}
\item \taskid{pick_up_the_alphabet_soup_and_place_it_in_the_basket_level3_sample1}
\item \taskid{pick_up_the_butter_and_place_it_in_the_basket_level1_sample2}
\item \taskid{pick_up_the_orange_juice_and_place_it_in_the_basket_level1_sample1}
\end{enumerate}

\paragraph{LIBERO-Goal.}
\textit{Robot Initial States}
\begin{enumerate}
\setlength{\itemsep}{0pt}
\item \taskid{open_the_middle_drawer_of_the_cabinet_view_0_0_100_0_0_initstate_11}
\item \taskid{open_the_top_drawer_and_put_the_bowl_inside_view_0_0_100_0_0_initstate_94}
\item \taskid{push_the_plate_to_the_front_of_the_stove_view_0_0_100_0_0_initstate_126}
\end{enumerate}
\Needspace{4\baselineskip}
\textit{Objects Layout}
\begin{enumerate}
\setlength{\itemsep}{0pt}
\item \taskid{open_the_middle_drawer_of_the_cabinet_moved_level1_sample1}
\item \taskid{open_the_top_drawer_and_put_the_bowl_inside_moved_level1_sample4}
\item \taskid{push_the_plate_to_the_front_of_the_stove_moved_level1_sample2}
\end{enumerate}

\paragraph{LIBERO-10.}
\textit{Robot Initial States}
\begin{enumerate}
\setlength{\itemsep}{0pt}
\item \taskid{KITCHEN_SCENE3_turn_on_the_stove_and_put_the_moka_pot_on_it_view_0_0_100_0_0_initstate_113}
\item \taskid{KITCHEN_SCENE4_put_the_black_bowl_in_the_bottom_drawer_of_the_cabinet_and_close_it_view_0_0_100_0_0_initstate_144}
\item \taskid{KITCHEN_SCENE6_put_the_yellow_and_white_mug_in_the_microwave_and_close_it_view_0_0_100_0_0_initstate_100}
\end{enumerate}
\Needspace{4\baselineskip}
\textit{Objects Layout}
\begin{enumerate}
\setlength{\itemsep}{0pt}
\item \taskid{KITCHEN_SCENE3_turn_on_the_stove_and_put_the_moka_pot_on_it_level1_sample4}
\item \taskid{KITCHEN_SCENE4_put_the_black_bowl_in_the_bottom_drawer_of_the_cabinet_and_close_it_level2_sample1}
\item \taskid{KITCHEN_SCENE6_put_the_yellow_and_white_mug_in_the_microwave_and_close_it_level1_sample2}
\end{enumerate}

\subsection{RoboCasa Kitchen}

Following the interaction-based grouping used by AAC~\cite{liang2026aac}, we categorize the 24 RoboCasa Kitchen tasks~\cite{nasiriany2024robocasa} into Relocation (10), Container (6), Rotation (5), and Button (3).
Each task is evaluated with 20 rollouts.

\paragraph{Relocation (10).}
\begin{enumerate}
\setlength{\itemsep}{0pt}
\item Counter to Cabinet (\texttt{PnPCounterToCab})
\item Counter to Stove (\texttt{PnPCounterToStove})
\item Sink to Counter (\texttt{PnPSinkToCounter})
\item Stove to Counter (\texttt{PnPStoveToCounter})
\item Counter to Sink (\texttt{PnPCounterToSink})
\item Counter to Microwave (\texttt{PnPCounterToMicrowave})
\item Cabinet to Counter (\texttt{PnPCabToCounter})
\item Microwave to Counter (\texttt{PnPMicrowaveToCounter})
\item Coffee Setup Mug (\texttt{CoffeeSetupMug})
\item Coffee Serve Mug (\texttt{CoffeeServeMug})
\end{enumerate}

\paragraph{Container (6).}
\begin{enumerate}
\setlength{\itemsep}{0pt}
\item Open Single Door (\texttt{OpenSingleDoor})
\item Close Single Door (\texttt{CloseSingleDoor})
\item Open Double Door (\texttt{OpenDoubleDoor})
\item Close Double Door (\texttt{CloseDoubleDoor})
\item Open Drawer (\texttt{OpenDrawer})
\item Close Drawer (\texttt{CloseDrawer})
\end{enumerate}

\paragraph{Rotation (5).}
\begin{enumerate}
\setlength{\itemsep}{0pt}
\item Turn Sink Spout (\texttt{TurnSinkSpout})
\item Turn On Sink Faucet (\texttt{TurnOnSinkFaucet})
\item Turn Off Sink Faucet (\texttt{TurnOffSinkFaucet})
\item Turn On Stove (\texttt{TurnOnStove})
\item Turn Off Stove (\texttt{TurnOffStove})
\end{enumerate}

\paragraph{Button (3).}
\begin{enumerate}
\setlength{\itemsep}{0pt}
\item Coffee Press Button (\texttt{CoffeePressButton})
\item Turn On Microwave (\texttt{TurnOnMicrowave})
\item Turn Off Microwave (\texttt{TurnOffMicrowave})
\end{enumerate}

\section{Additional Qualitative Results}

\begin{figure}[!htbp]
\centering
\includegraphics[width=\textwidth]{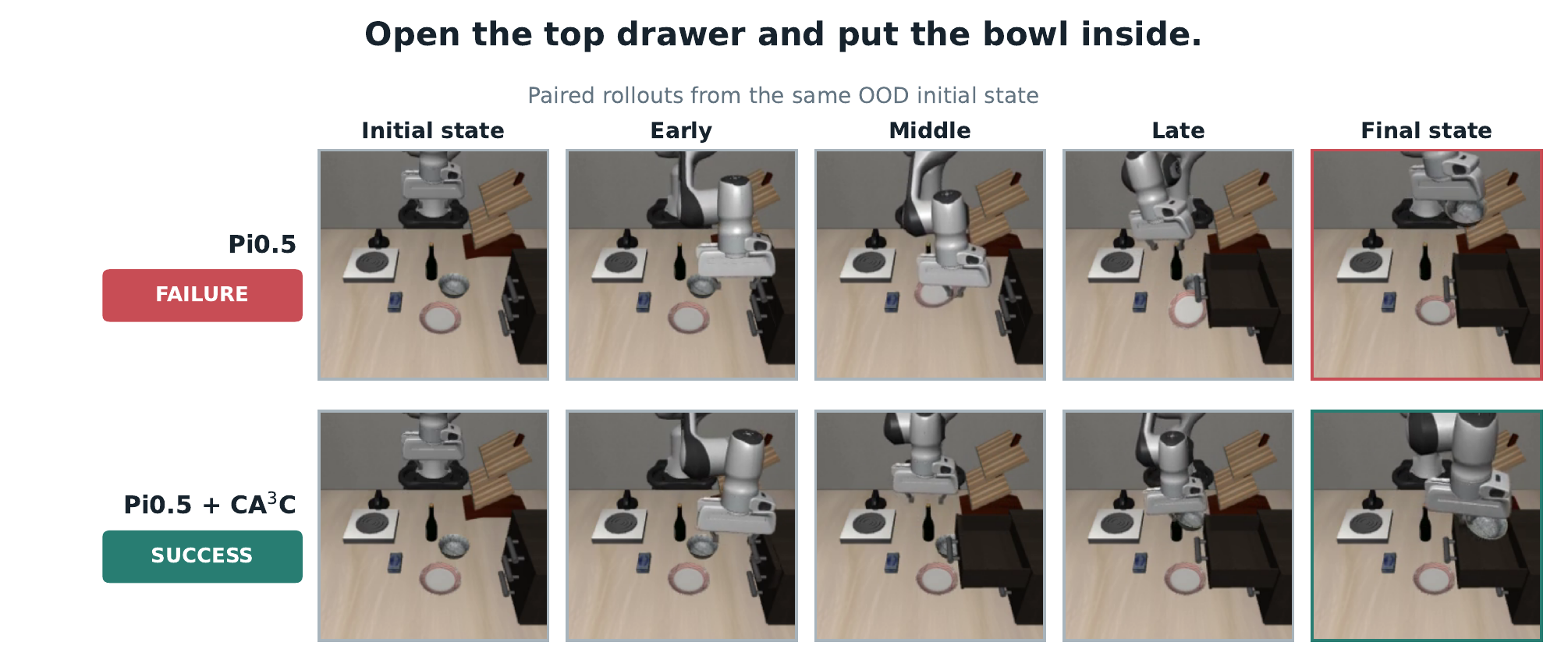}
\caption{Matched rollout comparison for the instruction \emph{Open the top drawer and put the bowl inside.}}
\label{fig:supp-top-drawer-bowl}
\end{figure}

\begin{figure}[!htbp]
\centering
\includegraphics[width=\textwidth]{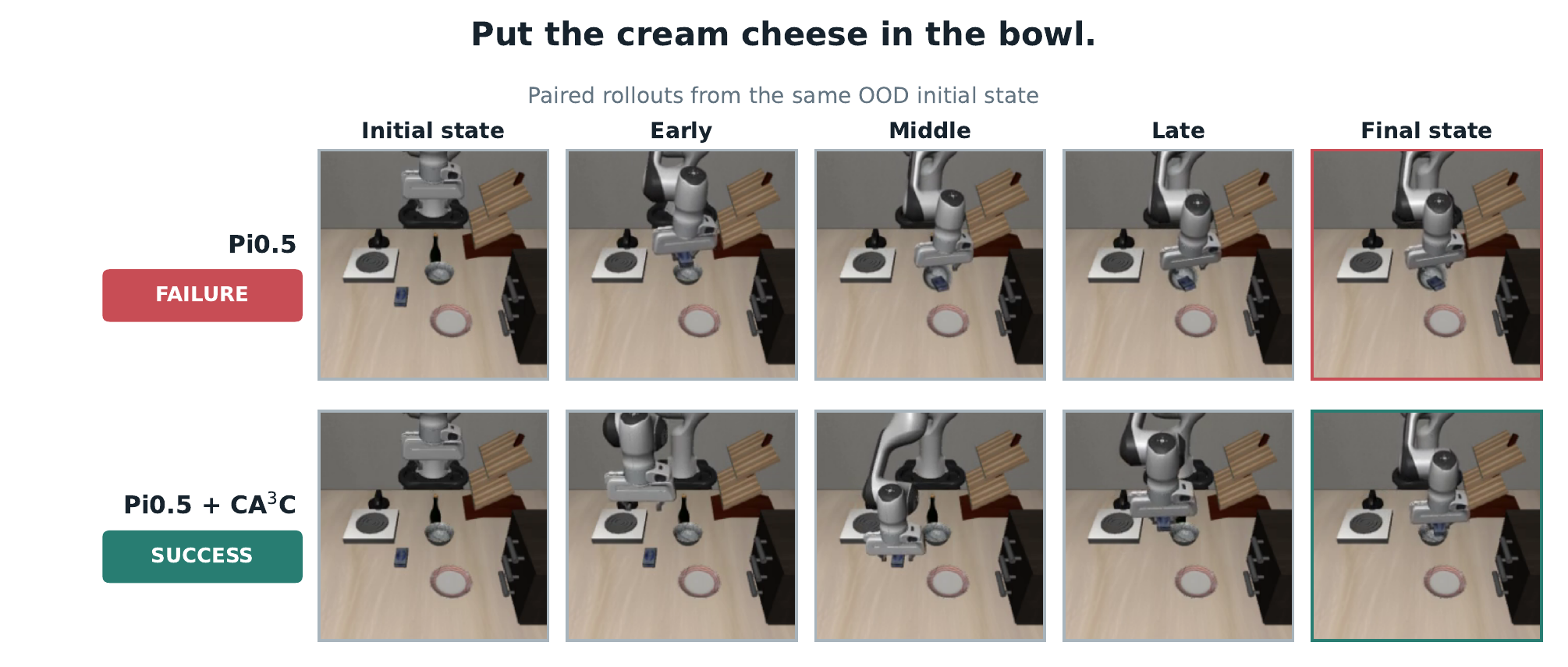}
\caption{Matched rollout comparison for the instruction \emph{Put the cream cheese in the bowl.}}
\label{fig:supp-cream-cheese-bowl}
\end{figure}

\begin{figure}[!htbp]
\centering
\includegraphics[width=\textwidth]{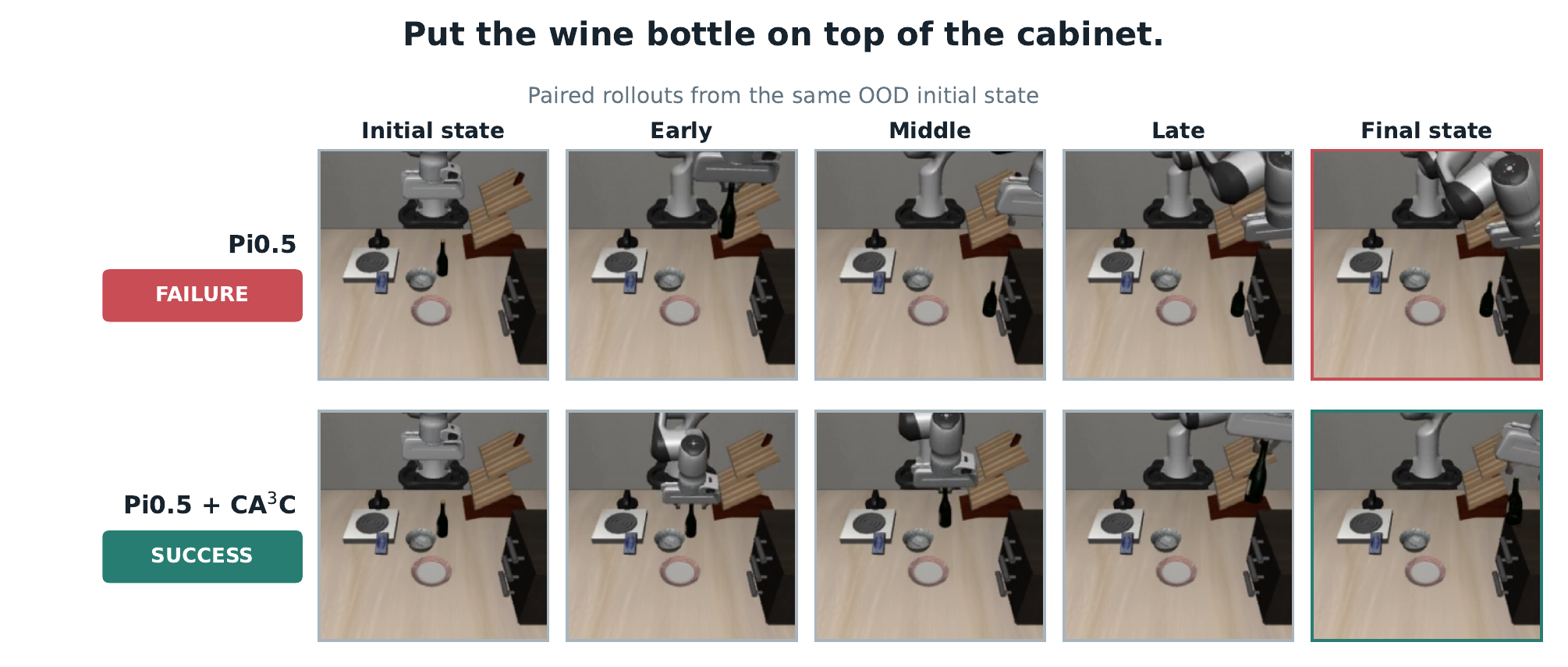}
\caption{Matched rollout comparison for the instruction \emph{Put the wine bottle on top of the cabinet.}}
\label{fig:supp-wine-bottle-cabinet}
\end{figure}

\begin{figure}[!htbp]
\centering
\includegraphics[width=\textwidth]{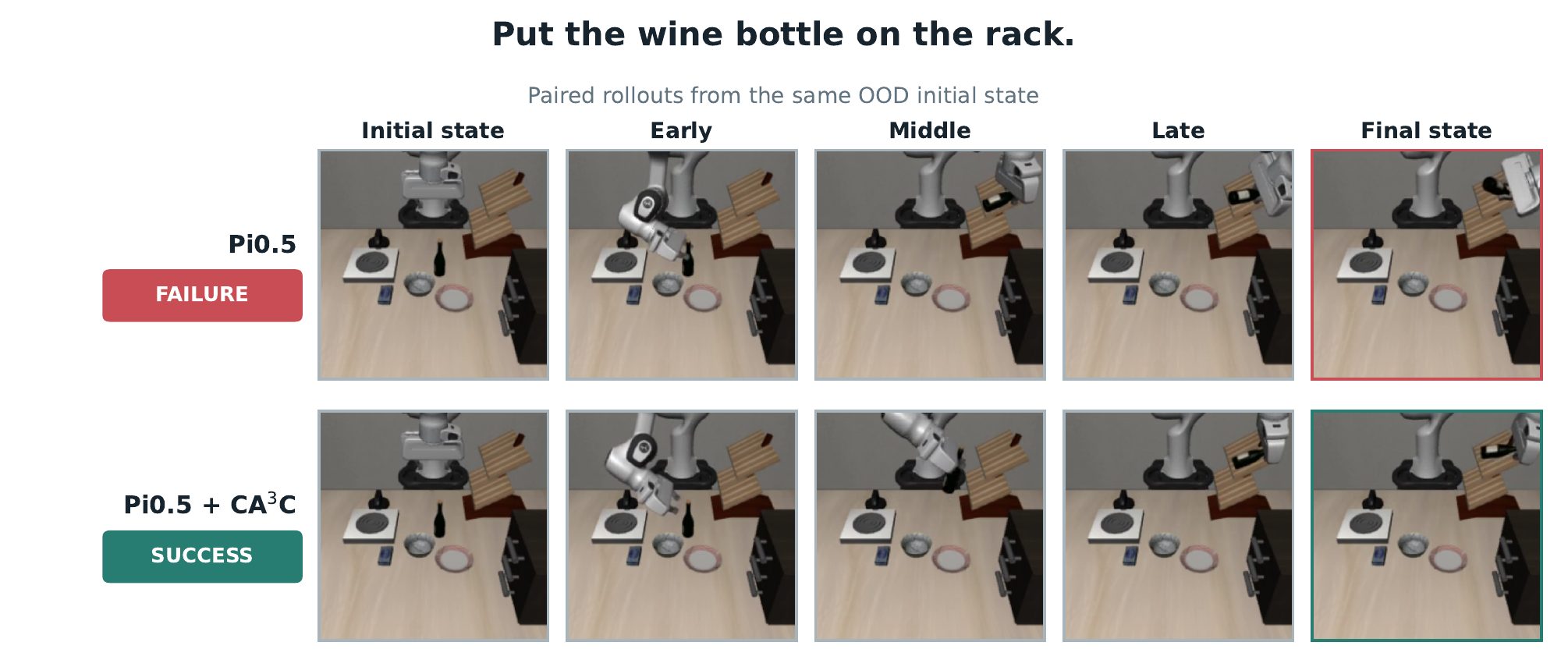}
\caption{Matched rollout comparison for the instruction \emph{Put the wine bottle on the rack.}}
\label{fig:supp-wine-bottle-rack}
\end{figure}

Figures~\ref{fig:supp-top-drawer-bowl}--\ref{fig:supp-wine-bottle-rack} provide additional matched rollouts on LIBERO OOD from identical initial states.
In each figure, the top row shows a failed rollout of $\pi_{0.5}$, while the bottom row shows a successful rollout of $\pi_{0.5}$+\method{}; red and green borders mark the respective final outcomes.

\end{document}